\documentclass[10pt,twocolumn,letterpaper]{article}

\usepackage{iccv}
\usepackage{times}
\usepackage{epsfig}
\usepackage{graphicx}
\usepackage{amsmath}
\usepackage{amssymb}

\usepackage[breaklinks=true,bookmarks=false]{hyperref}

\iccvfinalcopy % *** Uncomment this line for the final submission

\def\iccvPaperID{****} % *** Enter the ICCV Paper ID here

\begin{document}

%%%%%%%%% TITLE
\title{Beyond Forward Shortcuts:\\ 
	Fully Convolutional Master-Slave Networks (MSNets) with Backward Skip Connections for Semantic Segmentation}

\author{
	Abrar H. Abdulnabi\textsuperscript{1,2}\hspace{1cm}Stefan
	Winkler\textsuperscript{2}\hspace{1cm}Gang Wang\textsuperscript{3}
	%	\and
	%	Bing Shuai\textsuperscript{1}
	%	\and
	%	Gang Wang\textsuperscript{1}
	%	\and
	%	Stefan Winkler\textsuperscript{2}
	\vspace{3mm}\\
	\textsuperscript{1}~School of EEE, Nanyang Technological University (NTU), Singapore \\%{\tt\small \{abrarham001, bshuai001, wanggang\}@ntu.edu.sg}\\
	\textsuperscript{2}~Advanced Digital Sciences Center (ADSC), University of Illinois at Urbana-Champaign, Singapore \\ 	\textsuperscript{3}~Alibaba Group%{\tt\small \{abrarham001,%\\{\tt\small stefan.winkler@adsc.com.sg}}
}

\maketitle

	\begin{abstract}
	Recent deep CNNs contain forward shortcut connections; i.e. skip connections from low to high layers. Reusing features from lower layers that have higher resolution (location information) benefit higher layers to recover lost details and mitigate information degradation. However, during inference the lower layers do not know about high layer features, although they contain contextual high semantics that benefit low layers to adaptively extract informative features for later layers. In this paper, we study the influence of backward skip connections which are in the opposite direction to forward shortcuts, i.e.\ paths from high layers to low layers. To achieve this -- which indeed runs counter to the nature of feed-forward networks -- we propose a new fully convolutional model that consists of a pair of networks. A `Slave' network is dedicated to provide the backward connections from its top layers to the `Master' network's bottom layers. The Master network is used to produce the final label predictions. In our experiments we validate the proposed FCN model on ADE20K (ImageNet scene parsing), PASCAL-Context, and PASCAL VOC 2011 datasets.
\end{abstract}

\section{Introduction}\label{intro}
Deep Convolutional Neural Networks (CNNs) achieve impressive performance on many computer vision tasks, including image classification \cite{imagenet2012, zisserman2014, residual2015}, object detection \cite{fasterrcnn2015} and semantic segmentation \cite{fcn2015, deeplab2014, pyramid2016}. Deep CNNs extract hierarchical features by applying successive pooling and convolution layers with classifiers in an end-to-end trainable network \cite{viualize2013}. 
%Going deeper by adding more layers allow the network to variate and integrate richer low/mid/high features \cite{viualize2013, deeper2015}. However, the depth of the network obstructs the training process\cite{residual2015}. As a result, CNNs become substantially deeper overtime and more accurate but harder to train, yet, thanks to forward skip connections that make training easier \cite{residual2015, highway2015}. 

Fully Convolutional Networks (FCNs) \cite{fcn2015} are a natural extension of deep CNNs (e.g.  \cite{zisserman2014} is adapted for image segmentation). Unlike image classification, where only a single prediction is expected per image, the task is to generate a prediction label map for the whole input image. To maintain the original image resolution, FCNs employ upsampling layers to undo the effect of the downsampling produced by the pooling layers. Interestingly, FCNs contain skip connections from early pooling layers ($pool3$ and $pool4$) to upsampling layers. Such forward shortcut connections have two benefits: one is to avoid information degradation during the training process as evidenced by the residual block identity mapping in \cite{residual2015}; the other  is to integrate bottom layer features, which usually contain detailed low information for label prediction in the inference process.
%Skip connections create direct shorter paths to the supervisory error signal thus provide more efficient propagation and better training.
\begin{figure*}
	\centering
	\includegraphics[width=0.7\textwidth]{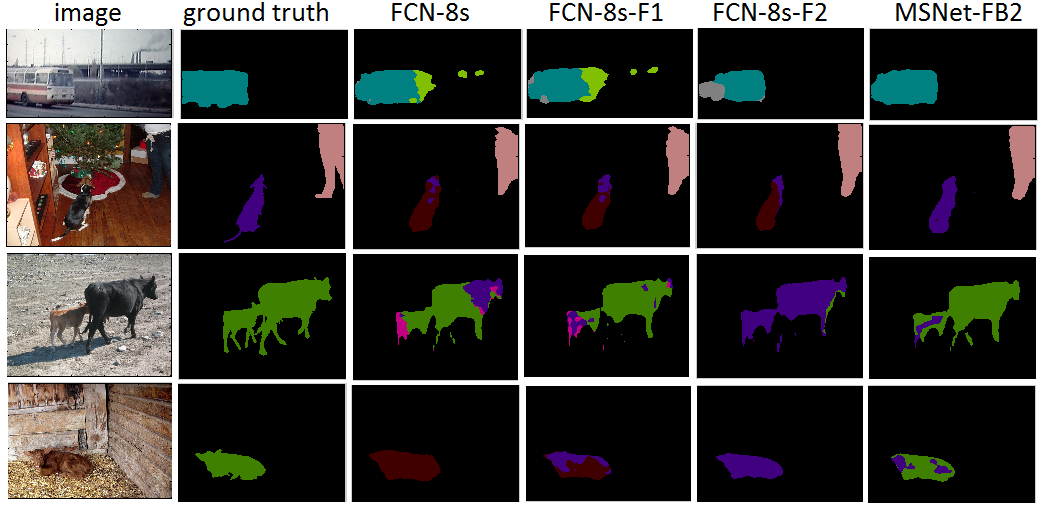}
	\caption{(Best viewed in color). Refer to Section \ref{baselines} for baseline definitions. Our MSNet model (last column) can correctly label object regions as the backward skip connections allow the lower layers to learn more informative features by receiving information from higher layers (contextually-aware). The sample images are from PASCAL VOC 2011 \cite{pascalvoc2011} validation set.}
	\label{Figure9}
\end{figure*}

In FCNs (or general deep CNNs), when extracting bottom layer features, the network is not aware of information at higher layers. This is actually a drawback: if bottom layers understand the context earlier, they can adaptively extract more informative features. For example, when labeling a car image region, once the bottom layers know the region contains a car, they may focus on the car-like visual patterns. Based on this motivation, we  propose backward skip connections which transfer information from higher layers to bottom layers directly. Figure \ref{Figure9} shows some interesting examples where our MSNet model (MSNet-FB2) is able to correctly label the object regions meanwhile other baselines fail. We can notice in these sample images that the overall high-level context plays an important role in discriminating the central object. Backward skip connections that are provided in our model help to pass such high-layer information as early as possible to low layers to extract informative features.   
% to exploit the very fine-grained details of bottom features and better recover the resolution while upsampling. In one hand, and in terms of contextual cues, these skip connections can be seen as providing multi-scale detailed contextual information for classification layers. 
%For example, Farabet \etal \cite{yanLecun2013} show that training multi-scale CNNs from raw pixels to extract dense feature vectors (encoding regions of multiple sizes centered on each pixel) does improve local classification of pixels. Other works with different design structures also show similar findings \cite{refineNet, pyramid2016, pixelNet}. On another hand, skip connections are indeed typical shortcut connections similar to the residual block identity mappings (where it skips between similar sized feature maps) \cite{residual2015}, but it skips between different blocks and apply convolutions to transform feature maps into similar dimensions. However, in our paper we still find that adding skip connections between similar sized feature maps still require a prior application of a simple $1\times1$ convolution to better transform the feature maps before aggregating them. 
\begin{figure*}
	\centering
	\includegraphics[width=0.75\textwidth]{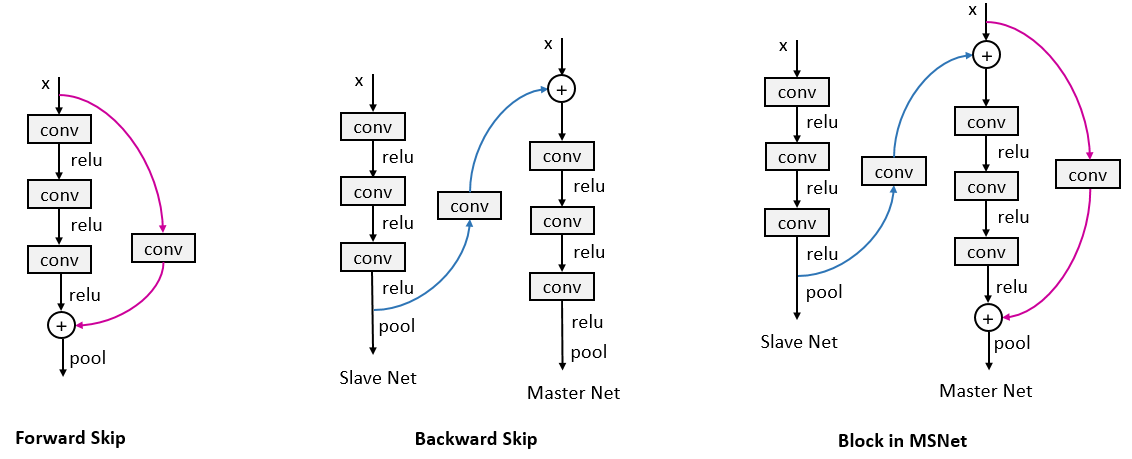}
	\caption{`Forward skip' is an illustration of one forward skip connection in our own baseline FCN-8s-F1. `Backward Skip' is our proposed design where we align two networks, a Master and Slave. The Slave network provides the backward skip connections from its higher layers to the Master network lower layers. `Block in MSNet' shows an illustration of our final model's skip connections, where additional forward skip connections have been added to the Master.}
	\label{Figure1}
\end{figure*}

%In semantic segmentation, global context (distant image regions) and the sub-regional/local context are both important for each pixel classification. Naturally, the higher layers contain more rich semantic meaning (high features) and less location information compared with the low layers. Thus, aggregating the local features in higher layers with lower layers' features allow these features in lower layers to be more contextually compatible and consistent on higher level.
Specifically, in our paper we observe the following: \textit{higher layers features are substantial to be conditioned on in the lower layers}. Directly skipping and fusing features  from the high layers back to lower layers shows consistently better results throughout our experiments. 

Given the fact that layers are stacked in a feed-forward manner in CNNs, connecting layers backward seems impossible as it would create cyclic loops inside the network. We overcome this problem by designing a pair of networks, each of which is originally an FCN-8s model. Only one of them -- the `Master' -- is used to produce label predictions. The other network -- the `Slave' -- is only responsible for providing the backward skip connections from its higher layers to the Master's lower layers.  The whole Master-Slave Network (MSNet) model is trained end-to-end. We evaluate our models on competitive benchmarks: ADE20K \cite{ade20k}, PASCAL-Context \cite{pascalcontext} and PASCAL VOC 2011 \cite{pascalvoc2011}, and we observe that our novel networks significantly outperform baselines which don't engage backward skip connections. 

\section{Related Work}\label{Related}
There are two research directions in the literature to tackle image segmentation, the first is by developing different CNN architectures which mainly evolve the idea of skip connections into different forms and designs, and the second is incorporating contextual modeling methods to collect different (multi-scale) contextual information that benefit local pixel classification.
\subsection{Evolution of Skip Connections in CNNs}
Highway Networks \cite{highway2015} propose to shortcut the flow through gating units to ease the training of a very deep CNN. ResNets \cite{residual2015} show competitive results by proposing to shortcut the flow through similar identity mapping connections. The basic idea is to skip from one layer (leaving some layers in between) and aggregate/sum with other later layer (usually by simple additive function). The skip connection can be gated/learnable or can be engaged directly (identity connection) and accompanied with prior convolutions, pooling or upsampling if needed. When FCNs are developed for segmentation \cite{fcn2015}, they propose skip connections that are designed to aggregate multi-scale feature maps of different channel dimensions. Similarly, simple convolutions are employed while skipping to further rematch feature dimensions prior fusing the feature maps. Beside, upsampling /bilinear interpolation is also employed to rematch feature spacial dimensions in case if the feature maps are being fused in different resolutions (transposed convolution can be also used instead). ResNets and Highway networks are considered to have shorter skips compared to FCNs which contain much longer skips (many layers are skipped in between) \cite{medicalskip}. Follow up works are developed to provide different styles of skips (both long and short) such as Stochastic Depth \cite{stochDepth} and Swapout \cite{SwapOut}. Meanwhile DenseNets are also proposed to have extremely dense forward skip connections (such as each layer is densely connected to all proceeding layers) \cite{denseNets, denseNetsSeg}. Another orthogonal line of works propose skips but in wider version of CNNs \cite{wideResNet, inception}. Different from all these forward skip connections, our model is designed with backward skip connections.

\subsection{CNNs and Contextual Modeling}
Engaging larger contextual information from other regions are usually important for local pixel classification. CNNs implicitly employ contextual information through its natural integration of multi-scale (hierarchical) features across the layers. However, excessively enlarging the receptive fields in the convolutions layers to capture longer/global context is insufficient \cite{CNNANDINTRARNN}. Skip connections in FCNs can be considered as one solution to overcome this limitation by aggregating the intermediate features for classification \cite{fcn2015}. Other works propose to explicitly encode global contextual information by stacking Recurrent Neural Network layers (RNNs) within CNNs. RNNs and their variants like Long Short-Term Memory networks (LSTMs) and Gated Recurrent Units (GRUs) are effective in encoding the dependency between different image regions and thus encode the global contextual information into local representations \cite{CNNANDINTRARNN, dagRNN}. Another line of works consider mainly the convolution operations and applied Atrous and dilated convolutions to perform contextual modeling \cite{dialatedCNN, deepLabJournal}. Conditional Random Fields (CRFs) and Markov Random Fields (MRFs) are also used to model the contextual label dependency \cite{deepLabJournal, CRFasRNN, effPiecCRF}. Most of the proposed methods stack CRFs on top of the classification layers in CNNs to serve as a post processing stage and can further capture the co-occurrence relationships between labels. However, our model is intended to improve the original FCNs structure and thus complementary to all these proposed models in the literature (e.g. our model can be easily extended and trained with RNNs or stacked with CRFs)

%\subsection{Our Feedback-like Inspired Design}
%Some earlier works are also similar, such as \cite{tu2008auto}, propose to utilize the output of classifiers as feedback to the next classification model. Similarly in \cite{pinheiro2014recurrent}, where feedback connections are added between the output and the input of CNNs in iterative manner (i.e. the output of the CNN in the previous iteration is fed back to the input of the same CNN in next iteration). Backward skip connections are indeed feedback signals that can steer the learning of better lower feature.  To the best of our knowledge, we are the first to propose backward skip connections within CNNs.
%These connections also encourage feature maps reuse and help to condition the bottom layers to learn features with higher semantics and abstractions for each pixel location. Thus, the features in lower layers are contextually consistent and compatible on higher level
\section{Fully Convolutional MSNets}\label{Framework}
While existing artificial CNNs have only forward connections, the biological brain -- for example in the visual system -- has complex connections of ample recurrent feedback between neurons \cite{bio1, bio2}. This inspired us to evolve FCNs to have variable length of  backward skip connections to mimic those feedback connections in the brain. 

The forward skip connections in basic FCN model encourage the reuse of feature maps from bottom layers to top layers and help to recover better resolution. In our model, we upgrade the FCN to contain backward skip connections from top layers to bottom layers. These connections enable the lower layers to be aware of higher-level information and context, as a result, the lower layers can extract features which are more informative for higher layers. In this section, we provide the details of our proposed model. First, we briefly review our first baseline which is `FCN-8s' structure \cite{fcn2015} since we adopt it as our basic building blocks in our MSNet model. Second, we further extend and improve FCN-8s to contain more forward skip connections and propose another two strong baselines. Third, we introduce our MSNet model that contains the backward skip connections only and discuss about the specifics of the proposed design. Finally, we further enhance our MSNet so that the Master network is upgraded into one of our improved baselines in addition to its backward skip connections. %However, in our final model design, we upgrade the Master network to contain more dense forward shortcuts than the Slave network, since its performance is more important and directly affect the final prediction maps. We could further upgrade the Slave network similarly but kept it light for economic reasons (saving memory and computations). We will discuss in detail about our model design specifics later.
\begin{figure*}
	\centering
	\includegraphics[width=0.9\textwidth]{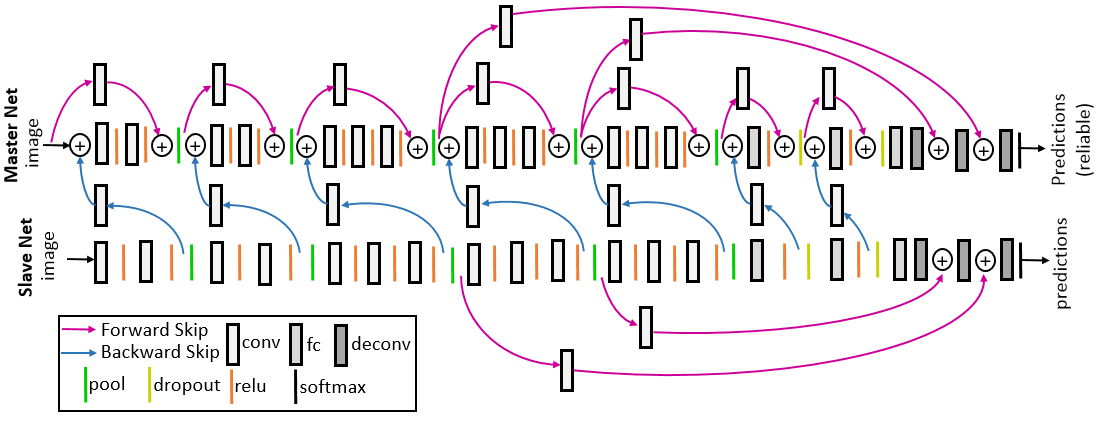}
	\caption{(Best viewed in colors). A detailed overview of our MSNet-FB1 model. The model consists of two networks; Master and Slave. Slave is a basic FCN-8s. The Master is a FCN-8s-F1  (as our baseline) and contains the backward skip connections taken from the Slave network. The forward propagation starts in the Slave network, and then the generated feature maps are fed back to lower layers in the Master network serving as backward skip connections within the Master. Notice that the backward skip connection (blue arrows) are exactly the inverse of the forward skip connections (purple arrows).}
	\label{Figure2}
\end{figure*}

\subsection{Baseline Models}
FCN originally are designed to extend image classification networks like the VGG Net \cite{zisserman2014} to tackle image segmentation. FCN architecture consists of: 1) a downsampling path that contains mainly seven convolutional blocks separated by five pooling layers (stacked in an alternating manner); 2) an upsampling path that contains upsampling layers to recover the original input resolution; 3) skip forward connections that are branched from the downsampling path and fused to the upsampling path to aid in recovering the lost spatial information and to produce detailed segmentation. The early design `FCN-32s' does not include skip connections, whereas FCN-16s has one skip connection from $pool4$ layer, and FCN-8s has skip connections from both $pool4$ and $pool3$ layers. Figure \ref{Figure2} shows the basic FCN-8s structure in the Slave network model. As shown in the downsampling path in the network, almost all the convolutional blocks contain 2-3 convolutions in addition to 2-3 non-linear activation layers (ReLU) inserted in between. Except for the last two convolutional blocks, they contain 3 fully-connected layers, 2 ReLUs and 2 dropout layers.

%\subsection{More forward skip connections in FCN}\label{forwards}
While FCN-8s only has two forward skip connections to the upsampling layers, we add extra forward skip connections to construct stronger baselines.
While designing the baseline structures, our main consideration is: \textbf{\textit{skipping from where to where and how many skips are adequate}}. We gradually add skips and shrink their lengths from low to high layers while recording the performance. Throughout the experimentation we observe that having at least one forward skip connection within the convolutional blocks and multiple skips from the direct preceding blocks provide most of the improvement gain in the performance (notice that blocks are separated by pooling or dropout layers). However, densely connecting each block to all preceding block is costly (as the network capacity will also grow), thus, the number of skip connections from neighboring blocks should be set properly.

We develop two stronger FCN-8s-based baselines with more forward skip connections. If the input to a convolutional block $L$ is $\textbf{x}_L$ (in our FCN-8s model, $ L = \{1,2,3,..,7\}$), the output from it is $\textbf{y}_{L}$, which becomes $\textbf{x}_{L+1}$ after a successive pooling or dropout layers. Mathematically, the general formula for adding $N$ forward skip connections from direct preceding blocks to block $L$ is:
\begin{equation} \label{forward_general_eq}
	\textbf{y}_{L} = \mathcal{F}(\textbf{x}_{L}) + \sum_{l = L-(N-1)}^{L} W_{l}\textbf{x}_{l}
\end{equation}
where $1 \leq N \leq L$ and $W_{l}$ is a convolutional transformation to rematch the feature dimensions\footnote[1]{In our experiments, we notice that adding a convolution layer to transform the feature map even if they originally have similar dimensions works better than directly aggregating them.}. $\mathcal{F}(.)$ is a convolutional block function that consists of multiple convolutions and non-linear activations (ReLU).

\textbf{FCN-8s-F1:}\label{baseline1} the design of our first baseline is a special case of Equation (\ref{forward_general_eq}) where $N=1$. In this case, each block has one skip connection within itself (from its beginning to its end) and the formula becomes:
\begin{equation} \label{F1_eq}
	\textbf{y}_{L} = \mathcal{F}(\textbf{x}_{L}) +  W_{L}\textbf{x}_{L}
\end{equation}

\textbf{FCN-8s-F2: }\label{baseline2} in the design of our second baseline, we add more skips from previous blocks, where we set $N=3$ and the formula becomes:
\begin{equation} \label{F2_eq}
	\textbf{y}_{L} = \mathcal{F}(\textbf{x}_{L}) + W_{L}\textbf{x}_{L} + W_{L-1}\textbf{x}_{L-1} + W_{L-2}\textbf{x}_{L-2}
\end{equation}
where $W_{L|L-1|L-2}$ are convolutional transformations (not shared). Notice that $W_{L-1|L-2}$ are accompanied with extra pooling layers beside the convolutions to rematch feature map spatial dimensions.

%Our final MSNet models contain either baseline designs in the Master network structure as illustrated in Figure \ref{Figure2} and Figure \ref{Figure4} respectively. The baselines are the same as the Master network but without any backward skip connections. In summary, we find the best performing basic structure is to always contain skip connections from the beginning of each convolutional block to the end of it. Adding dense forward connections from different block is beneficial but not always. We only consider the downsampling path of the FCN. Notice that in our second baseline we skip after applying the pooling on the output of the convolutional blocks.

\subsection{Master-Slave Network Design}\label{msnet}
In this paper, we propose to add backward skip connections to FCNs. However, if we simply add them in an FCN, we will create cyclic loops in the network, thus, training becomes a problem. To overcome this challenge, we propose a Master-Slave network design.

Figure \ref{Figure2} shows in details our proposed final structure. The model consists of two FCN-8s networks (\textit{none} of their parameters are shared). Only one network which is the Master is considered to produce reliable final predictions. The other network which is the Slave is only responsible to feed the Master with the required backward skip connections. The Slave is not provided with any backward skip connections from the Master network and its predictions are totally ignored. However, both networks are optimized to minimize similar losses. 

Notice that the backward skip connections are the inverse of our forward skip connections as previously formulated in Equation (\ref{forward_general_eq}). Hence,  the general formula for adding $P$ backward skip connections to block $L$ of the Master network is denoted as: 
\begin{equation} \label{backward_general_eq}
	\textbf{y}_{L}^{m} = \mathcal{F}(\textbf{x}_{L}^{m} + \sum_{l=L}^{L+(P-1)} U_{l} \textbf{y}_{l}^{s})
\end{equation}
where $\textbf{y}_{L}^{m}$ is the output of block $L$ in the Master network and $\textbf{y}_{L}^{s}$ is the output of block $L$ in the Slave network ($m$ and $s$ refer to Master and Slave). $U_{l}$ is a convolutional transformation applied on the feature maps from the Slave network.

Based on Equation (\ref{backward_general_eq}), our Master network can learn to discover visual patterns from $\textbf{x}_{L}^{m}$ that are more compatible with $\textbf{y}_{L}^{s}$.

In our model designs, we define two structures as counterparts to our previously designed baselines. \textbf{MSNet-B1} is the inverse of FCN-8s-F1 while \textbf{MSNet-B2} is the inverse of FCN-8s-F2. In MSNet-B1, each block in the Master network is provided with one backward skip from the Slave (from the end of the peer block in the Slave). In MSNet-B2, each block in the Master network is provided with multiple backward skip connections from proceeding blocks in the Slave network (we set $P=3$).

We further upgrade our Master network to also have internal dense forward skip connections similar to our baselines. Overall, the forward and backward skip connections in block $L$ in the Master network are formulated as:
\begin{equation} \label{general_eq}
	\textbf{y}_{L}^{m} = \mathcal{F}(\textbf{x}_{L}^{m} + \sum_{l=L}^{L+(P-1)} U_{l} \textbf{y}_{l}^{s}) + \sum_{l = L-(N-1)}^{L} W_{l}\textbf{x}_{l}^{m}
\end{equation}
where $W_{l}$ and $U_{l}$ are convolutional transformations.

\begin{figure}
	\centering
	\includegraphics[width=0.5\textwidth]{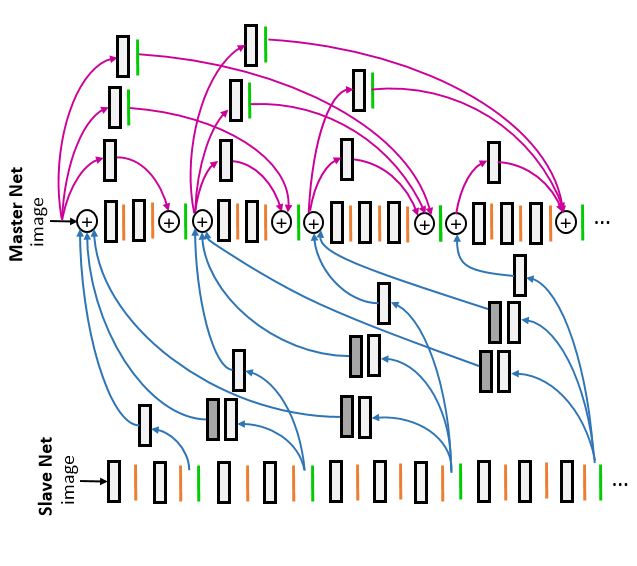}
	\caption{(Best viewed in colors). An illustration of several convolutional blocks of our MSNet-FB2 with dense skip connections. Notice that the Master network is FCN-8s-F2. We set $P=3$ and $N=3$. Here the dense backward skip connections are in inverse directions to the dense forward skip connections. Refer to Figure \ref{Figure2} for the Legend.}
	\label{Figure4}
\end{figure}

\begin{table*}
	\begin{center}
		\begin{tabular}{l|c|c|c|c|c|c|c|c|c}
			\hline
			Algorithm &  \multicolumn{3}{c}{ADE20K} & \multicolumn{3}{c}{PASCAL-Context} & \multicolumn{3}{c}{PASCAL VOC 2011}\\	\hline
			{}   & PA(\%)  & CA(\%)  & IU & PA(\%)   & CA(\%) & IU(\%)  & PA(\%)  & CA(\%) & IU(\%)  \\
			FCN-8s & 71.3 & 40.3 & 29.4 & 67.5 & 52.3 & 39.1 & 91.2 & 77.6 & 65.5 \\ 
			\hline
			FCN-8s-F1 & 73.5 & 41.2 & 31.4 & 70.7 & 52.3 & 40.7 & 91.1 & 76.4 & 66.0 \\
			MSNet-B1 & 75.1 & 42.7 & 32.9 & 71.8 & 52.5 & 41.7 & 92.1 & 77.6 & 68.3 \\
			\hline
			MSNet-FB1 & 74.8 & 41.7 & 32.5 & \textbf{73.3} & \textbf{53.5} & \textbf{42.1} & 91.9 & 77.2 & 67.7 \\
			\hline
			FCN-8s-F2 & 73.7 & 42.0 & 31.7 & 69.0 & 51.1 & 38.8 & 91.9 & 77.5 & 66.9 \\
			MSNet-B2 & 75.2 & 43.0 & 33.3 & 71.9 & 50.6 & 39.7 & 92.0 & 78.0 & 68.6 \\
			\hline
			MSNet-FB2 & \textbf{75.7} & \textbf{43.8} & \textbf{33.8} & 70.1 & 50.3 & 39.9 & \textbf{92.3} & \textbf{80.3} & \textbf{69.2} \\
			\hline
		\end{tabular}
	\end{center}
	\caption{Results on the ADE20K \cite{ade20k} (validation set), PASCAL-Context \cite{pascalcontext} and PASCAL VOC 2011 (validation set) \cite{pascalvoc2011}. Refer to Section \ref{baselines} for detailed definitions.\label{tab:t1}}
\end{table*}
In summary, our model mainly contains three types of components in the downsampling path: 1) convolutional block; 2) forward-skip-fuse (skip forward through a convolution layer and fuse in a summation layer); 3) backward-skip-fuse (skip backward through a convolution layer and a summation layer). Forward-skip-fuse may contain an extra pooling layer if the fusion is between multi-scale feature maps. Similarly, backward-skip-fuse may have an upsampling layer to properly fuse the multi-scale feature maps. Details of these components are shown in Figure \ref{Figure4}. The Figure visualize an example with $L=3$, $N=3$ and $P=3$ in MSNet-FB2 model.
%	\begin{equation}
%		\textbf{y}_{L}^{m} = \mathcal{F}(\textbf{x}_{L}^{m} + U_{L}\textbf{y}_{L}^{s} + U_{L+1} \textbf{y}_{L+1}^{s}) + W_{L}\textbf{x}_{L}^{m} + W_{L-1}\textbf{x}_{L-1}^{m}
%	\end{equation}
%	where $W_{L|L-1}$ and $U_{L|L+1s}$ are convolutional transformations in the Master and Slave networks.

\section{Experiments}\label{Experiments}
We evaluate our method on three competitive image segmentation benchmark datasets: ADE20K, PASCAL-Context and PASCAL VOC 2011. We provide evaluation details of our proposed network and other hyperparameters and design choices.
\subsection{Datasets}
\textbf{ADE20K \cite{ade20k}} has 20210 training, 2000 validation and 3352 testing images. It is created for the ImageNet scene parsing challenge. The segmentation task in this dataset is to label each pixel to one of the 150 semantic classes. 
%Statistically, this dataset has imbalanced class distribution. We found that applying class balancing helps similar to \cite{ContextDriven14CVPR, dagRNN}.

\textbf{PASCAL-Context \cite{pascalcontext}} has 4998 training images and 5105 testing images. The images are sampled originally from the PASCAL VOC 2010 dataset and re-labeled at the pixel-level for the segmentation task. In total, there are 540 classes, however, in our experiments we only consider the task of labeling 59 classes (most frequent classes) for evaluation.

\textbf{PASCAL VOC 2011 \cite{pascalvoc2011}} have around 11287 training images, 736 images as validation set and 1111 testing images. The task here is to classify each pixel into one of 21 categories, including 20 foreground object classes and one background class.

\textbf{Evaluation} We use three main metrics to evaluate our method following the recent literature: the percentage of all correctly classified pixels - Pixel Accuracy (\textbf{PA}), Per-class Accuracy (\textbf{CA}), and the Intersection-Over-Union (\textbf{IU}).

\subsection{Baselines and Methods} \label{baselines}
In summary, our baselines and models are: 

\textbf{FCN-8s}: the basic FCN structure with two forward skip connections from $pool3$ and $pool4$.

\textbf{FCN-8s-F1}: improved FCN-8s structure that contains more forward skip connections within each convolutional block (refer to Section \ref{baseline1} for details).

\textbf{FCN-8s-F2}: densely connected FCN-8s structure that contains $N=3$ forward skip connections (refer to Section \ref{baseline2} for details).

\textbf{MSNet-B1}: a Master-Slave network model where the backward skip connections from the Slave network to the Master network are in inverse direction to the forward skip connections in FCN-8s-F1. Here, the Master and Slave networks are basic FCN-8s models.

\textbf{MSNet-B2}: similar to MSNet-B1, but the backward skip connections from the Slave network to the Master network are in inverse direction to the forward skip connections proposed in baseline FCN-8s-F2. The number of the backward skips are set to $P=3$.

\textbf{MSNet-FB1}: similar to MSNet-B1, but the Master is upgraded to be our baseline FCN-8s-F1. The Slave network is basic FCN-8s.

\textbf{MSNet-FB2}: similar to MSNet-B2, but the Master is upgraded to be our baseline FCN-8s-F2. The Slave network is basic FCN-8s.

\subsection{Optimization Details}
MSNet is end-to-end differentiable and trainable using the backpropagation algorithm (we use the stochastic gradient descent (SGD) with momentum). Both the Master and the Slave networks are trained to minimize the cross entropy log-loss averaged over all image patches (we only consider the semantically labeled ones), but they are \textit{not} jointly trained. In the forward pass, the propagation begins in the Slave network. When it reaches the first backward skip connections, it triggers the propagation in the Master network to start. The execution order is alternating between both networks to make sure that all the feature maps of the backward skip connections are generated prior to any forward propagation in a lower layer in the Master network. In the backward pass, the execution order is reversed.

Hyperparameters like the momentum is initialized and fixed to $0.9$, meanwhile the learning rate is initialized to $10^{-5}$, and it decays exponentially with the rate of 10\% after $10$ epochs, and after $5$ epochs when it finish the first 30 epochs. We follow \cite{fcn2015} to train our basic FCN-8s models. We initialize all our networks with the VGG ImageNet-pretrained model \cite{zisserman2014}. We used the publicly available MatConvNet MATLAB implementations \cite{vedaldi15matconvnet}. The reported results are based on the model trained for 50 epochs. After we generate the class likelihood per each pixel in the image, we calculate the cross entropy loss for training. Thus, the error signal of each image is averaged across all valid/semantically-labeled pixels (we ignore the contribution of unlabeled pixels in the loss calculation). We also report the learning and inference run time of our baselines and our methods as shown in Table \ref{tab:t5}. 
\begin{table}
	\footnotesize
	\begin{center}
		\begin{tabular}{l|c|c}
			\hline
			Algorithm   & $\sim$Train (ms) & $\sim$Test (ms)\\
			\hline
			FCN-8s  & $183.4$ & $55.6$ \\
			FCN-8s-F1 & $195.4$ & $54.6$ \\
			FCN-8s-F2 & $238.4$ & $65.6$ \\
			MSNet-B1 & $388.3$ & $108.7$ \\
			MSNet-B2 & $411.7$ & $114.7$ \\
			MSNet-FB1 & $409.2$ & $113.4$ \\
			MSNet-FB2& $475.4$ & $130.2$ \\
			\hline
		\end{tabular}
	\end{center}
	\caption{We report training and testing run time on the validation set of ADE20K (averaged over 20 trials for an input image of $384\times384$ on NVIDIA GeForce GTX Titan X and the companion CPU is Intel Xeon E5-2650 v2 at $2.60$GHz).\label{tab:t5}}
\end{table}

\begin{table}
	\footnotesize
	\begin{center}
		\begin{tabular}{|l|c|c|c|c|}
			\hline
			Algorithm &   \multicolumn{3}{c}{Validation} & Test \\
			{}   & PA(\%)  & CA(\%)  & IU(\%)  & IU(\%) \\\hline
			MSNet-FB1 & 91.9 & 77.2 & 67.7 & 69.3 \\ 
			MSNet-FB1+CRF & 92.4 & 78.3 & 68.9 & 70.3 \\
			MSNet-FB2 & 92.3 & 80.0 & 69.2 & 70.5 \\
			MSNet-FB2+CRF& \textbf{92.8} & \textbf{80.9} & \textbf{70.3} & \textbf{71.4}\\
			\hline
		\end{tabular}
	\end{center}
	\caption{Results on the validation and testing sets of PASCAL VOC 2011 \cite{pascalvoc2011}. MSNet-*+CRF refers to applying fully-connected CRF on top of the label predictions generated from our MSNet models (post processing step).\label{tab:t6}}
\end{table}
\subsection{Results}
Table \ref{tab:t1} show the prediction results on the ADE20K, PASCAL-Context and PASCAL VOC 2011 datasets.

\textbf{Adding Backward Shortcuts helps:} our model MSNet-B1 improves over FCN-8s by 3.5\%, 2.6\% and 2.8\% respectively (IU). While MSNet-B2 improves over FCN-8s by 3.9\%, 0.6\% and 3.1\% respectively (IU). The experiments show that the higher layers' features are indeed important during lower layers' features extraction. Higher abstractions can help lower layers to learn more informative lower features (as high-level context is explored early). 

\begin{figure*}
	\centering
	\includegraphics[width=0.6\textwidth]{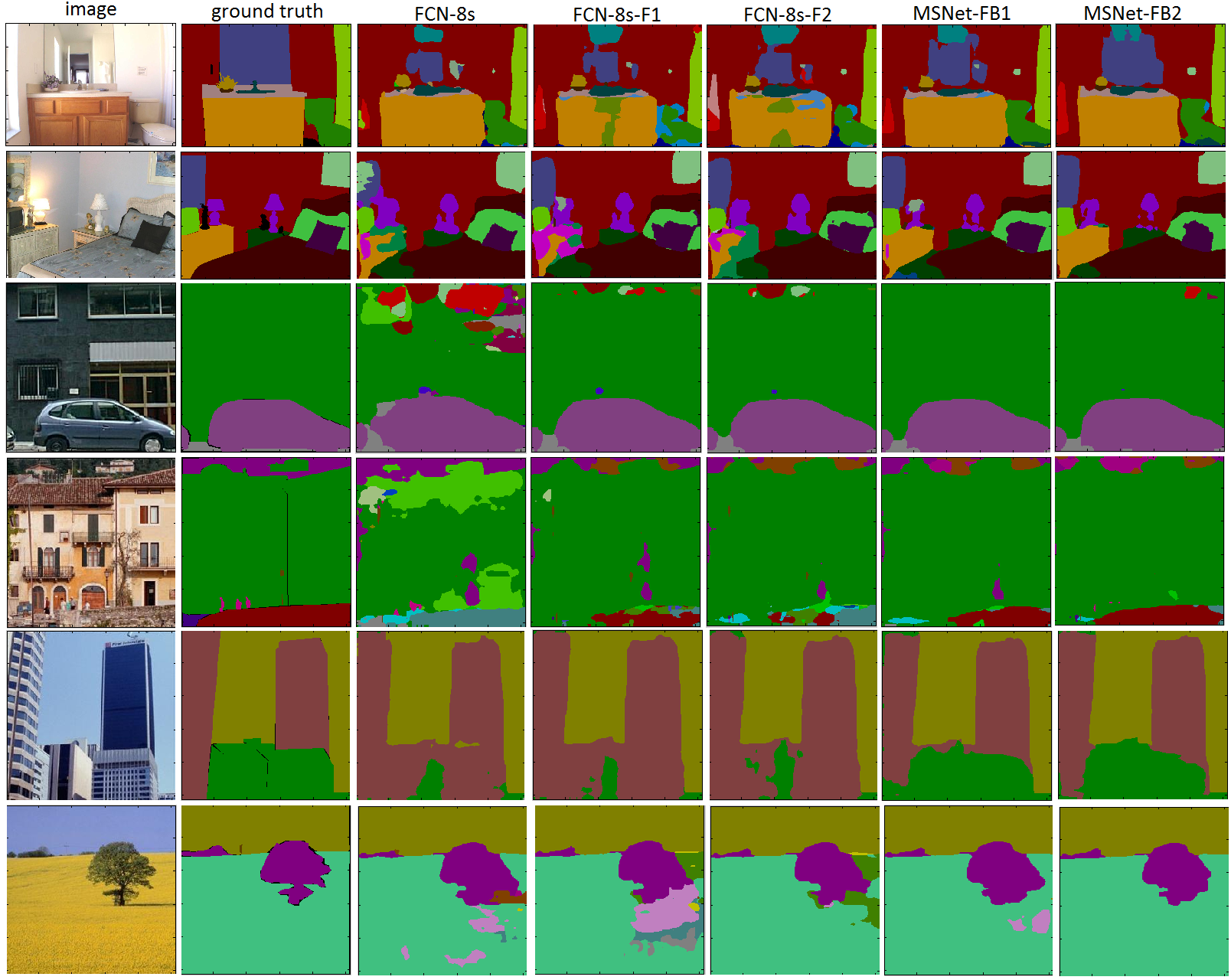}
	\includegraphics[width=0.6\textwidth]{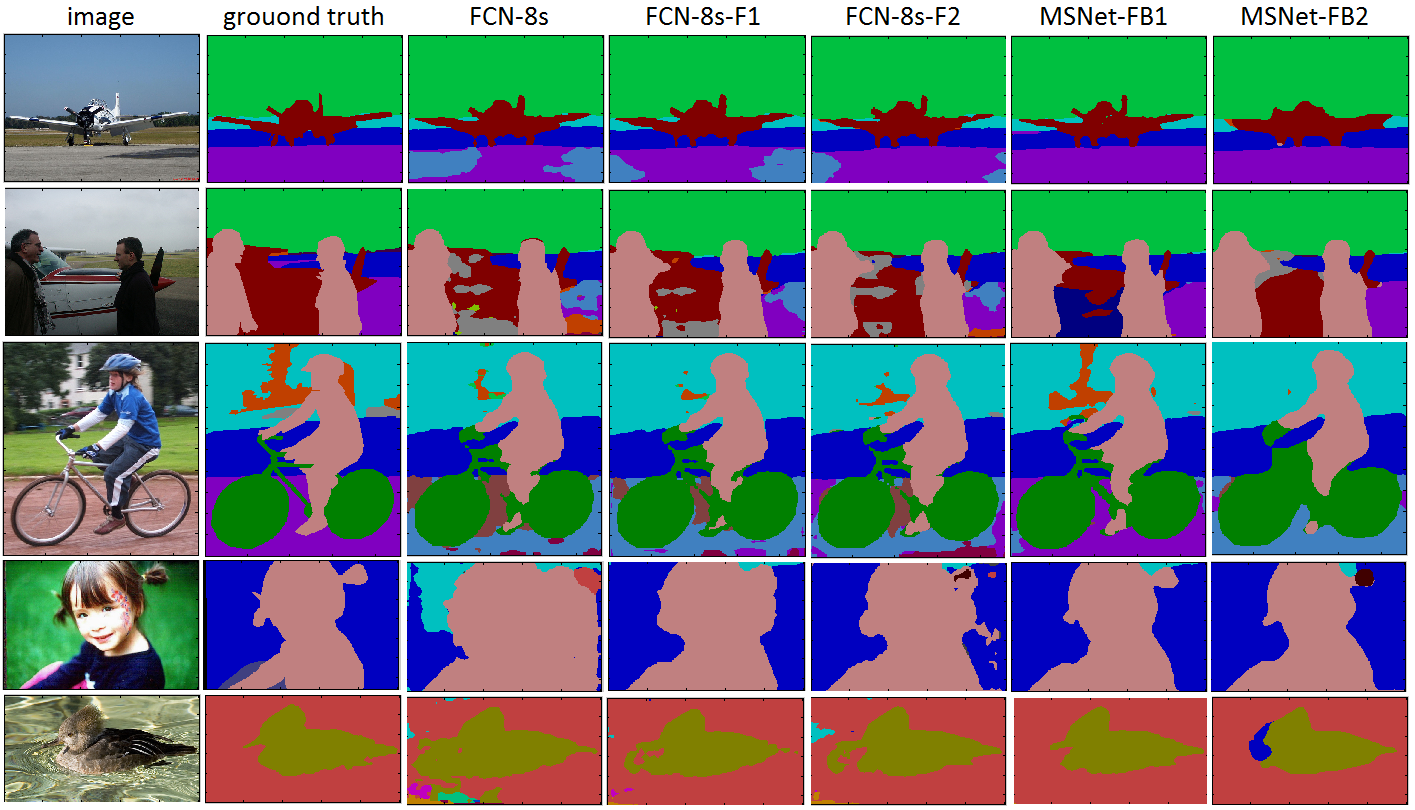}
	\caption{(Best to be zoomed in and viewed in colors) Examples of some qualitative labeling results. For each row, we show the input image, its ground truth, label prediction maps for FCN-8s, FCN-8s-F1, FCN-8s-F2 and our proposed models MSNet-FB1 and MSNet-FB2. The test images are from ADE20K validation set \cite{ade20k} (top) and PASCAL-Context \cite{pascalcontext} (bottom).The MSNet-FB1 and MSNet-FB2 have extra backward skip connections and improve the quality of the label prediction maps compared with baselines. The FCN-F2 and FCN-F2 baselines improve over FCN-8s as they have more forward skip connections.}
	\label{Figure8}
\end{figure*}

\textbf{Backward and Forward Shortcuts are complementary:} as shown in Table \ref{tab:t1}, MSNet-FB1 improves over FCN-8s-F1 by 1.1\%, 1.4\% and 1.7\% respectively (IU). While MSNet-FB2 improves over FCN-8s-F2 by 2.1\%, 1.1\% and 2.3\% respectively (IU). The results show that both types of skip connections are complementary and important. We notice the MSNet-FB2 performs better than MSNet-FB1 on PASCAL VOC and ADE20K, but worse on PASCAL Context. This may be due to that MSNet-FB2 is more dense and complex (higher capacity) and thus require more training data to generalize better. For example, ADE20K contains the highest number of training examples and MSNet-FB2 performs well on it.

\textbf{Backward Shortcuts outperform Forward Shortcuts:} interestingly, MSNet-B2 which contains only backward skip connections outperforms both baselines FCN-8s-F1 and FCN-8s-F2 in all datasets. Similarly, MSNet-B1 consistently outperform its counterpart FCN-8s-F1. These findings show that high layers' features are indeed beneficial for extracting more informative low features.

\textbf{MSNets are complementary to CRFs:} In order to evaluate the orthogonality of our MSNet with other common methods like CRF, we perform post-processing on top of MSNet using the Fully-connected Conditional Random Fields implementations \cite {fullyConectedCRFOrg}. Fully-connected CRF is one of the most common contextual modeling methods applied in scene labeling (it is one of the state-of-the-art works that applied in \cite{deeplab2014, deepLabJournal}). As shown in Table \ref{tab:t6}, MSNet-FB2+CRF and MSNet-FB1-CRF outperform MSNet-FB2 and MSNet-FB1 by at least 1.0\% respectively. This performance gain illustrates that our proposed method is complementary to other existing frameworks.

\textbf{Qualitative Results of MSNets:} we show the qualitative accuracy results in Figure \ref{Figure8}. The quality of the label predictions is significantly improved over FCN-8s as more forward skip connections are engaged (the baselines) and also when backward skip connections are engaged (MSNet models).

\textbf{Comparison with the State-of-the-art:} we compare the performance of our MSNet models with other state-of-the-art methods on image segmentation. Notice that we didn't engage any additional training data (like from Microsoft Common Objects in Context (MS COCO) dataset \cite{coco}). We also didn't employ any Batch Normalization layers as our main focus is to improve the original design of FCNs. It is also important to mention that there are some methods that use different settings than us like adopting very deep Residual Networks \cite{residual2015} (e.g. \cite{pyramid2016}, \cite{VeryDeep} and \cite{deepLabJournal}) or employing Dilated convolutions \cite{deepLabJournal}. However, we use VGG-based networks, and note our backward skip techniques can be potentially used as well to enhance their models. The quantitative results comparison on ADE20K, PASCAL-Context and PASCAL VOC 2011 datasets are listed in Table \ref{tab:t2}, \ref{tab:t3} and \ref{tab:t4} respectively. Notice that in Table \ref{tab:t3}, we compare on the validation set of ADE20K. For Table \ref{tab:t4}, we evaluate on PASCAL VOC 2011 test set. Our models perform competitively on all datasets to those models which have similar sittings with us. 

\begin{table}
	\footnotesize
	\begin{center}
		\begin{tabular}{|l|c|c|c|}
			\hline
			Algorithm & PA(\%)   & CA(\%)  & IU(\%) \\
			\hline
			FCN-8s \cite{fcn2015} & 65.9 & 46.5 & 35.1  \\
			FCN-8s \cite{fcn2016}  & 67.5 & 52.3 & 39.1 \\    
			DeepLab \cite{deepLabJournal} & - & - & 37.6  \\
			DeepLab-CRF \cite{deepLabJournal} & - & - & 39.6\\
			HO-CRF \cite{hoCRF}& - & - & 41.3\\
			CNN-CRF \cite{effPiecCRF} &  71.5 & \textbf{53.9} & \textbf{43.3}\\
			CRF-RNN \cite{CRFasRNN} & - & - & 39.3\\
			ParseNet \cite{parseNet} & 67.5 & 52.3 & 39.1\\
			ConvPP-8 \cite{convPP} & - & - & 41.0\\
			PixelNet \cite{pixelNet}& - & 51.5& 41.4\\
			O2P \cite{o2p} & - & - & 18.1\\
			CFM \cite{cfm}& - & - & 34.4\\
			BoxSup \cite{boxSup} & - & - & 40.5\\
			\hline
			MSNet-FB1 &  \textbf{73.0} & 53.5 & 42.1\\
			\hline
		\end{tabular}
	\end{center}
	\caption{Performance comparison on PASCAL-Context \cite{pascalcontext}.\label{tab:t3}}
\end{table}
\begin{table}
	\begin{center}
		\begin{tabular}{|l|c|c|c|}
			\hline
			Algorithm & PA(\%) & CA(\%) & IU(\%) \\
			\hline
			SegNet \cite{segNet} & 71.0 & 31.1& 21.6\\
			FCN-8s \cite{fcn2015} & 71.3 & 40.3& 29.4\\
			DilatedNet \cite{dialatedCNN} & 73.6 &  44.6& 32.3\\
			Cascade-SegNet \cite{ade20k} & 71.8 & 37.9& 27.5\\
			Cascade-DilatedNet \cite{ade20k}& 74.5 & \textbf{45.4} & \textbf{34.9} \\
			\hline
			MSNet-FB2 &  \textbf{75.7} & 43.8 & 33.8\\
			\hline
		\end{tabular}
	\end{center}
	\caption{Performance comparison on the validation set from ADE20K \cite{ade20k}.\label{tab:t2}}
\end{table}
\begin{table}
	\begin{center}
		\begin{tabular}{|l|c|}
			\hline
			Algorithm   & IU(\%) \\
			\hline
			BerkeleyRC \cite{berkdec} & 39.1\\
			SDS \cite{sds} &  52.6\\
			R-CNN \cite{richRCNN} &  47.9\\
			FCN-8s \cite{fcn2015}  & 62.7\\
			FCN-8s \cite{fcn2016}  & 67.5\\
			Zoomout \cite{zoomout} & 64.4\\
			CRF-RNN\cite{CRFasRNN} & \textbf{72.4} \\
			\hline
			MSNet-FB2 & 70.5\\
			\hline
		\end{tabular}
	\end{center}
	\caption{Average performance comparison on the test set from PASCAL VOC 2011 \cite{pascalvoc2011}.\label{tab:t4}}
\end{table}
\section{Conclusions}\label{Conclusions}
In this paper, a new type of skip connection is presented. We upgraded the structure of FCN-8s with two designs: first we add more forward skip connections from lower layers to higher layers (beside its original two skips from $pool3$ and $pool4$), and second we propose new backward skip connections that fuse higher layers feature maps into lower layers. In order to achieve this, we design our model as a pair of Master and Slave networks. They are twin versions of FCN-8s, however, the Master is upgraded with more forward skips. The Slave is only responsible for feeding the Master with backward skip connections from its higher layers to the Master lower layers. The Master is accountable to produce the final prediction maps. Our experiments show that lower layers can adaptively extract more informative and beneficial features if they are aware as early as possible with high layer feature representations.

{\small
	\bibliographystyle{ieee}
	\bibliography{egbib}
}
\end{document}